\documentclass{article}

\usepackage{PRIMEarxiv}

\usepackage[utf8]{inputenc} 
\usepackage[T1]{fontenc}    
\usepackage{hyperref}       
\usepackage{url}            
\usepackage{booktabs}       
\usepackage{amsfonts}       
\usepackage{nicefrac}       
\usepackage{microtype}      
\usepackage{lipsum}
\usepackage{fancyhdr}       
\usepackage{graphicx}       
\usepackage{amsmath}

\title{Signed Distance Field based Segmentation and Statistical Shape Modelling of the Left Atrial Appendage
}

\author{
  Kristine Aavild Juhl, Jakob Slipsager, Rasmus Paulsen \\
  DTU Compute \\
  Technical University of Denmark \\
  Kongens Lyngby, Denmark \\
  \texttt{\{kajul, jmsl, rapa\}@dtu.dk} \\
   \And
  Oscar Camara \\
  Physense  \\
  Universitat Pompeu Fabra
  \\ Barcelona, Spain \\
  \texttt{oscar.camara@upf.edu} \\
   \AND
  Ole de Backer, Klaus Kofoed \\
  Department of Cardiology\\
  Rigshospitalet \\ 
  Copenhagen, Denmark \\
  \texttt{\{ole.de.backer, Klaus.Kofoed\}@regionh.dk} \\
}

\begin{document}
\maketitle

\begin{abstract}
    Patients with atrial fibrillation have a 5-7 fold increased risk of having an ischemic stroke. 
In these cases, the most common site of thrombus localization is inside the left atrial appendage (LAA) and studies have shown a correlation between the LAA shape and the risk of ischemic stroke. 
These studies make use of manual measurement and qualitative assessment of shape and are therefore prone to large inter-observer discrepancies, which may explain the contradictions between the conclusions in different studies. 
We argue that quantitative shape descriptors are necessary to robustly characterize LAA morphology and relate to other functional parameters and stroke risk. 

Deep Learning methods are becoming standardly available for segmenting cardiovascular structures from high resolution images such as computed tomography (CT), but only few have been tested for LAA segmentation. 
Furthermore, the majority of segmentation algorithms produces non-smooth 3D models that are not ideal for further processing, such as statistical shape analysis or computational fluid modelling.  
In this paper we present a fully automatic pipeline for image segmentation, mesh model creation and statistical shape modelling of the LAA. 
The LAA anatomy is implicitly represented as a signed distance field (SDF), which is directly regressed from the CT image using Deep Learning. 
The SDF is further used for registering the LAA shapes to a common template and build a statistical shape model (SSM). 
Based on 106 automatically segmented LAAs, the built SSM reveals that the LAA shape can be quantified using approximately 5 PCA modes and allows the identification of two distinct shape clusters corresponding to the so-called chicken-wing and non-chicken-wing morphologies. 

\end{abstract}


\section{Introduction}
\begin{figure}[t!]
\centering
\includegraphics[width=.6\linewidth]{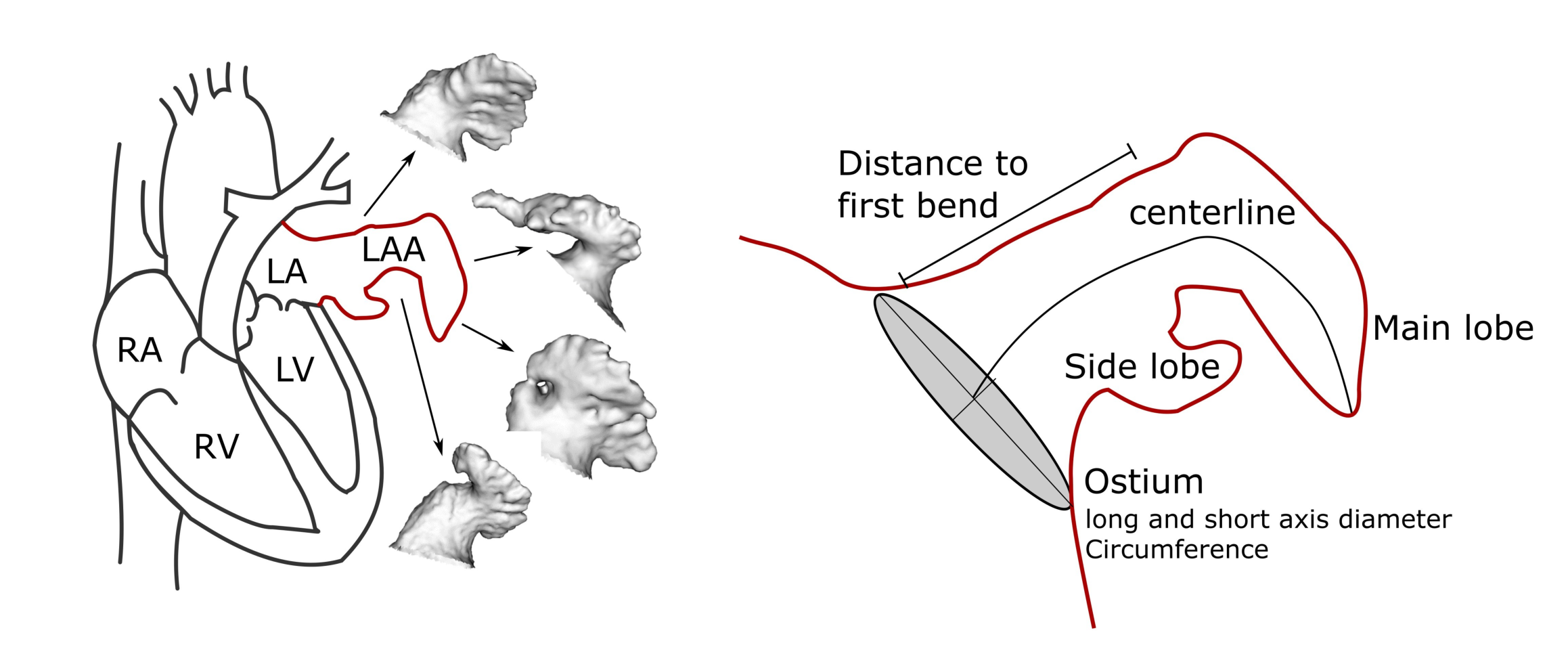}
\caption{A) Schematic illustration of the human heart together with four examples of Left Atrial Appendage (LAA) morphologies from different subjects in our data set. RA: right atrium, RV: right ventricle, LA: left atrium, LV: left ventricle. 
B) Examples of commonly used measures of LAA morphology}
\label{fig:LAAintro}
\end{figure}

Patients suffering from atrial fibrillation have a 5-7 times higher risk of experiencing an ischemic stroke, and in these patients the most common thrombus location site is inside the LAA, where up to 99\% of the reported thrombi are located \cite{Glikson2020, Cresti2019}. 
The left atrial appendage (LAA) is a complex tubular structure originating from the left atrium (LA) and is known to have large morphological variability. 
Figure \ref{fig:LAAintro} shows a schematic illustration of the human heart together with four examples of LAA morphologies from different subjects.

Independently of thrombotic risk factors, blood flow velocities inside the LAA are predictors of clot formation, for which reason the size and shape of the LAA are hypothesized to correlate with thrombogenic risk and thereby also stroke risk. 
Several clinical studies have aimed to answer the question of how LAA morphology and stroke risk is correlated, but the conclusions from the studies are not in convergence.  
An increased stroke risk  has been linked to short LAA length, small orifice and extensive trabeculation in the LAA \cite{Khurram2013}. 
But also with increased orifice size and decreased flow velocities in the LAA \cite{lee2017additional}, and  to the so-called non-chicken-wing morphologies \cite{di2012does} (A morphology without an obvious bend or folding back of the LAA anatomy on itself).
The contradictions and inconclusiveness of these studies may come from the lack of well-defined morphological parameters and standardized measurement protocols. 
Instead, more robust, stable and automatic methods to extract LAA anatomy and derive quantitative morphological parameters are needed.

Computed Tomography (CT) is in many clinical sites becoming the modality of choice when imaging the LAA, and with high-resolution images comes the possibility of doing high-quality segmentations. 
Only few have attempted to segment the LAA from CT images. 
Prior to the Deep Learning era, researchers proposed classical image analysis techniques such as deformable models with mesh inflation techniques \cite{Grasland2010combination}, graph-cut algorithms \cite{Wang2017} and adding spheres of decreasing radii along a regressed centerline through the LAA \cite{Leventic2019occluderplacement}.
More recently, Convolutional Neural Networks (CNNs) and Reinforcement Learning has been adapted for the LAA segmentation task. 
For instance, a 2D CNN was trained with adversarial learning \cite{Takayashiki2019} or in combination with a 3D conditional random field \cite{Jin2018}. 
Reinforcement Learning was used to localize seeds for a geodesic distance segmentation algorithm \cite{AlWalid2018}. 
Neither of these methods make full use of the 3D information as it can be done using 3D convolutions. Furthermore their aim is to accurately localize voxels in the image that belong to the LAA with no considerations to how the segmentation can used for further analysis. 

In this paper we aim to go beyond voxel-wise segmentation and wish to use the segmentation outputs for building patient-specific mesh models. 
Such models can be used to run computational fluid simulations to estimate in-silico haemodynamics \cite{Aguado2019, Mill2020} or to conduct statistical shape analysis \cite{slipsager2018statistical} as we demonstrate in this paper. 
It is an ongoing research topic, to investigate how triangulated meshes can be represented in a Deep Learning framework, but standard operations such as convolution, pooling, etc. are still lacking compared to the Euclidean domain. 
An alternative to representing the anatomy as a mesh or binary voxel-grid is to use implicit functions, as for example a Signed Distance Field (SDF). 
The SDF can be evaluated on a voxel-grid and thereby allows one to make use of voxel-based methods for processing. 
\cite{Park2019} and \cite{Kleineberg2020} demonstrates how a surface can be represented as an SDF in a deep learning framework to do mesh reconstruction, completion and generation. 
In medical image segmentation the SDFs has been used for regularization, either as a part of the loss function \cite{caliva2019distance} or as an auxiliary task during training \cite{dangi2019distance, bui2019multi, Wang2017, xue2019shape}). 
Besides using the SDF, other medical segmentation algorithms aims to incorporate contour loss to encoruage smooth segmentation results \cite{Abdeltawab2020,Gu2020}.

Once an LAA mesh model is available, metrics can be obtained to characterize the LAA morphology both quantitatively (ostium perimeter, distance to first bend, etc.) \cite{Korsholm2020} or qualitative (grouping into chicken-wing, cactus, cauliflower and windsock) \cite{wang2010left}. 
In this work, we aim to move away from manually defined and measured morphological features and instead propose to build a statistical shape model (SSM) from the collection of 3D LAA objects to represent the average shape and the variations hereof \cite{Ambellan2019}. 
The technique has been widely used to describe both normal and pathological variations in different anatomies \cite{Krason2019, Ma2019}, but has not been utilized to describe the highly variable LAA morphology before.
Building an SSM requires a correspondence mapping between the individual LAAs and one way of achieving this is by registering all shapes to a common template \cite{paulsen2002building}. 

In this paper we present a fully automatic pipeline to segment the LA with the LAA from CT images, register it to a common template and systematically investigate and describe the variations in LAA shape. 
The pipeline contains a Deep Learning method for predicting SDFs directly from CT images. 
The zero-level isosurface in the predicted SDF represents a smooth and continuous 3D mesh model of the LA, which can be used for further analysis. 
The predicted SDF can also be thresholded to produce an accurate segmentation map with dice-scores on par with state-of-the-art LAA segmentation methods. 
We follow up the segmentation method with describing how point correspondence is obtained between 106 complex LAA shapes, which is utilized to build an SSM that can be used to investigate and describe the LAA morphology in a quantitative manner. All code and trained models is publicly available\footnote{\url{https://github.com/kristineaajuhl/LAAmeshing_and_SSM}}.

\section{Materials and methods}
\begin{figure*}[!t]
\centering
\includegraphics[width=0.9\textwidth]{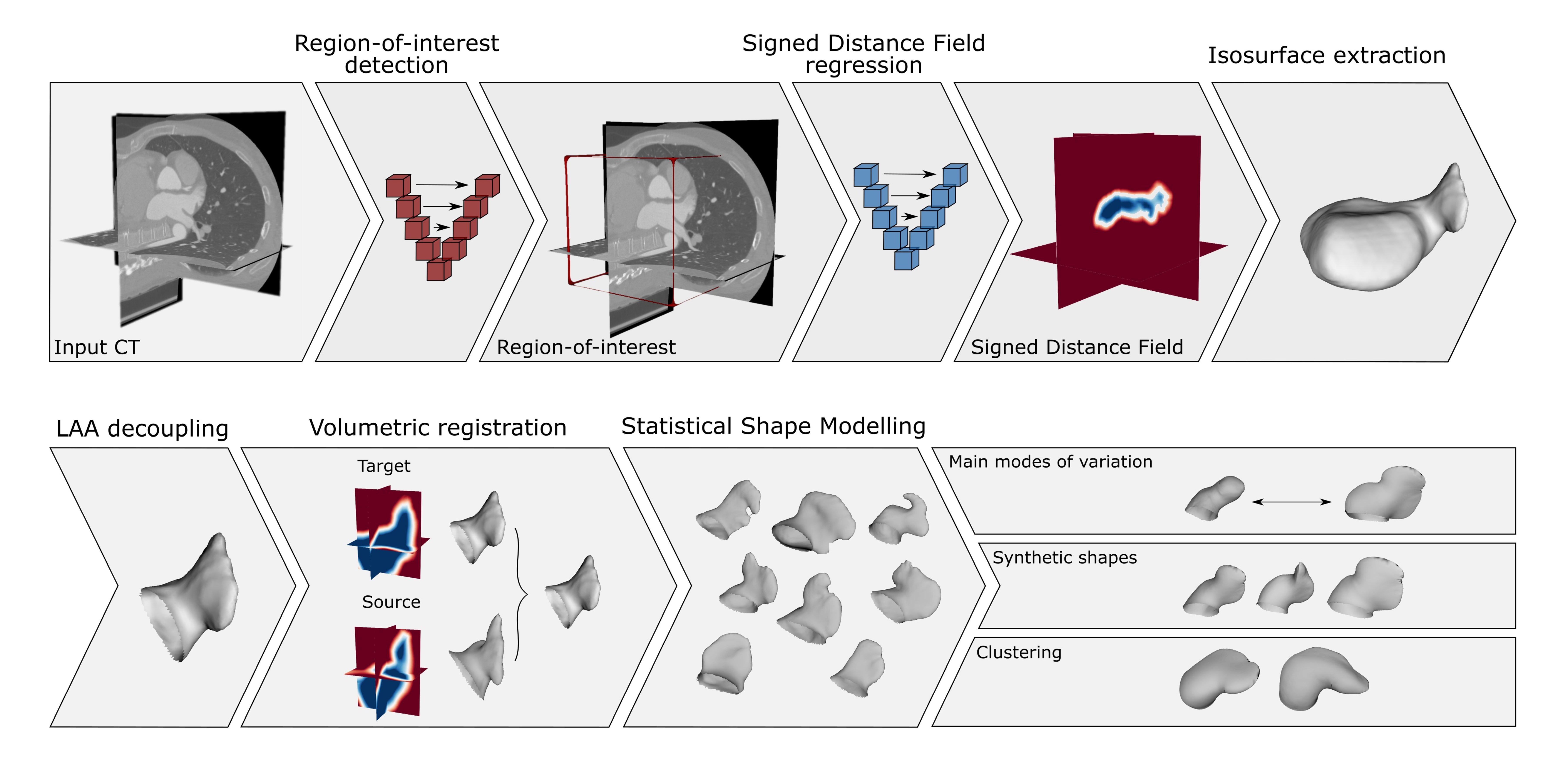}
\caption{Overview of the full processing pipeline. A region-of-interest is detected around the left atrium (LA) and a signed distance field (SDF) is regressed based on it. The zero-level isosurface is extracted and the left atrial appendage (LAA) is decoupled from the LA. All LAA shapes are registered to a common template (Source) using volumetric registration and a statistical shape model (SSM) is built. The SSM enables us to investigate the main modes of variation, generate realistic synthetic shapes and conduct unsupervised clustering.}
\label{fig:methods}
\end{figure*}

We propose a method for learning SDFs directly from CT images and describe how SDFs are further used for registering all shapes to a common template. 
The registered shapes are used to build an SSM investigating the LAA morphology. 
The full pipeline can be seen in Figure \ref{fig:methods}.
The top part visualizes the process from input CT-image, through region-of-interest (ROI) detection to SDF regression and isosurface extraction. 
Afterwards the surface is registered to a common template, the LAA is decoupled from the LA and the point correspondence is refined over the LAA only. 
The LAA shapes in correspondence is used to build a SSM which can be utilized to investigate the main modes of variation, create new synthetic shapes and cluster the LAAs into similar shapes.

\subsection{Segmentation and mesh model creation}\label{sec:segmentation}
\subsection*{Region of Interest (ROI)}
Since the LA only fills a small part of the acquired CT volume, a region-of-interest (ROI) was extracted around the LA and LAA. 
We downsampled the full input image to the processing size of $64^3$ voxels and predicted a preliminary low-resolution segmentation of the LA using a 3D U-net.
A 3D bounding box with a side length of 140 mm was centered at the center-of-mass of the largest connected component in the segmentation output. 
The region was resampled isotropically to the input size of the second network. 

\subsection*{Predicting SDFs: Voxel Wise Regression (VWR)}
The goal of the proposed method is to regress an SDF directly from the ROI of the CT-scan.
An SDF is an implicit representation of a surface, where the magnitude of a point in the SDF represents the distance to the nearest point on the surface, and the sign indicates whether the point is inside (-) or outside (+) the surface. 
The surface is thus directly retrievable from the zero-level-set. 
The practical application of SDFs was pioneered by the seminal work on surface reconstruction from point clouds \cite{Hoppe1992surface}. 
Here a sampled SDF was created based on an unorganized point cloud and the approximating surface was extracted as the zero-level isosurface. 
Our work can be seen as an extension of this idea, where the SDF is estimated directly from raw medical scans using CNNs.

In typical image segmentation with CNNs the aim is usually to do binary classification of each voxel, deciding whether it belongs to the anatomy or not. 
We denote this Voxel-Wise Classification (VWC). 
In our approach, we instead aim to regress the signed distance from the center of each voxel to the nearest point on the surface of interest. 
This requires regressing a continuous value in each voxel, and we denote this Voxel-Wise Regression (VWR)

The feature extraction from the CT image is done with a 3D U-net architecture.
To achieve the regression abilities, we replace the final softmax layer with a linear activation function.
We choose a linear activation function above non-linear activation functions, since the values around zero are the most important and small changes in the features leading into this layer should thus not change the value too much.

The loss function used for regression is a weighted mean squared error (wMSE) loss given as:
\begin{equation}\label{eq:wMSE}
  \text{wMSE}(y_i, \hat{y}_i) = \frac{1}{N} \sum_i^N \frac{1}{|y_i| + \lambda} \cdot |y_i - \hat{y}_i|^2 ,
\end{equation}

\noindent where $y$ is the true distance field, $\hat{y}$ is the predicted distance field, N is the number of voxels and $\lambda$ is a weighting parameter ensuring the stability of the loss at values close to zero, being set to $\lambda = 0.001$ by experimental validation.
The MSE term allows us to do regression in each voxel, whereas the weighting ensures that most attention is given to values near the zero-level isosurface, where the final boundary will be placed.

The surface is found as the zero-level isosurface in the predicted SDF at the processing size of $64^3$ voxels. 
The surface extraction method keeps track of the true voxel size and the dimensions of the extracted surface are therefore directly comparable to the original input images. 
The label map is found by linear upsampling of the SDF to the original resolution of the input image and thresholding at zero. 

\subsection*{Baseline: Voxel-Wise Classification (VWC)}
We compare our results to standard segmentation using voxel-wise classification (VWC). 
We keep the network architecture and training parameters as similar to the VWR network as possible, to ensure fair comparison. 
The final layer uses soft-max activation and the network is optimized using binary cross entropy loss. 
In our proposed method of SDF regression the loss function is weighted inversely by the distance field. 
We therefore also compare against a similar inverse weighting of the binary cross entropy loss used for voxel-wise classification and denote it SDF regularized VWC. 
For both methods the output probability map is linearly upsampled to the original resolution, where it is thresholded at 0.5 for the labelmap and the 0.5-level isosurface is extracted using a marching cubes algorithm. 

\subsection*{Implementation details}
We used the same backbone network architecture for both ROI and SDF regression. 
The backbone network is a 3D U-net \cite{cciccek20163d} with padding at each layer, transposed convolutions and spatial dropout.
All networks are trained in PyTorch \cite{pytorch} using Adam Optimization, a learning rate of $10^{-4}$ and 5\% dropout.
63 images are used for training, 7 images for validation to employ early stopping and 20 images are used exclusively for testing and evaluation. 
The network is trained on a single Nvidia GeForce GTX 1080 Ti with a batch size of 1. 

We predict SDFs in a supervised manner, which requires SDFs that capture the true 3D nature of the surface for each training example. 
To create the ground truth SDFs, we first extract the 0.5-level isosurface of manually annotated training images using a marching cubes algorithm \cite{lorensen1987marching} and smooth the surfaces using three iterations of Laplacian smoothing with a relaxation factor of 0.1. 
The Laplacian smoothing evens the artefacts from the voxelization of the image. 
The SDFs are created on a uniform voxel volume, where each voxel value represents the signed distance from its center to the nearest oriented point on the surface. 
For more details and a ready-to-use implementation, the reader is referred to \cite{paulsen2009markov}. 
The method ensures 3D coherent SDFs, but does not make mathematically true SDFs (ie. gradient of zero in all points). 
The SDFs is truncated at $\pm5$ mm.

\subsection{Statistical Shape Modelling}\label{sec:ssm}
\begin{figure*}[!t]
\centering
\includegraphics[width=\textwidth]{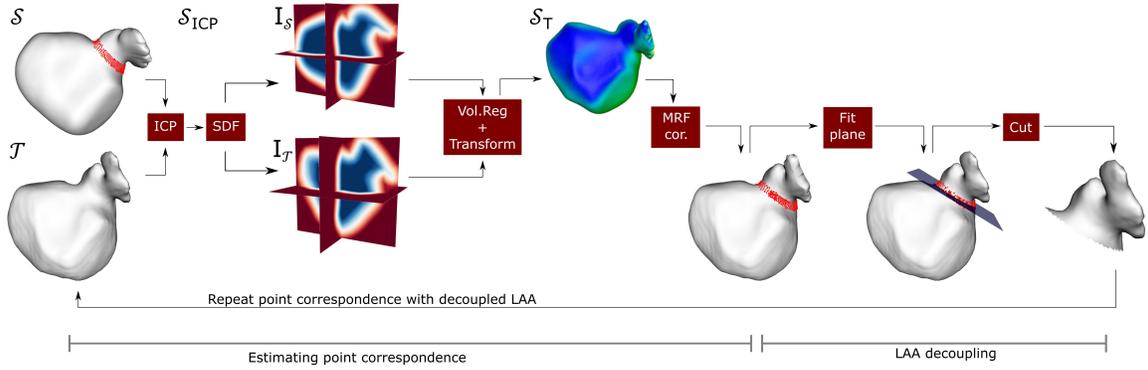}
\caption{Point correspondence is established between the source and each target using volumetric registration (Vol.Reg). The two surfaces are pre-aligned using an iterative closest point (ICP) algorithm, before the SDFs are registered and the source is transformed with a B-spline transformation. All points on the transformed source are projected to the target mesh using markov random field regularization of the correspondence field (MRF cor.). Based on the neck points defined on the original source, we find the best fitting plane that decouples the LAA from the LA. The method for estimating point correspondence is repeated, this time using the decoupled LAAs for Source and Target.}
\label{fig:registration}
\end{figure*}

Statistical shape modelling is a technique, where the variation of shapes across a population can be modeled using principal component analysis (PCA).
In the following sections we describe how each LAA surface is registered to a common template and how this is used for decoupling the LAA from the LA and further for investigating morphological variations of the LAA. 
The whole process is illustrated in Figure \ref{fig:registration}.

\subsection*{Template registration} \label{sec:registration}
An SDF-based volumetric registration approach has shown to be effective when registering complex surfaces \cite{paulsen2017a} and we base our surface registration on that framework. 

We denote the template or source surface $\mathcal{S}$ and register it to each of the target surfaces $\mathcal{T}$ in the training set. 
Initially we randomly pick one surface from the training data as the source. 
After registering this source to all other surfaces in the training set, a new template is created as the mean of the Procrustes aligned outputs \cite{dryden1998a}. 
We repeat the process for three iterations, each time using the new template as the source.
After three iterations we found that the source surface was no longer dependent on which example that was chosen in the first iteration. 

The source $\mathcal{S}$ is aligned to one target $\mathcal{T}$ with an iterative closest point (ICP) algorithm using a similarity transform. 
We now denote the source surface $\mathcal{S}_{ICP}$. 

We use a non-rigid volumetric registration algorithm to register $\mathcal{S}_{ICP}$ to $\mathcal{T}$.
We convert the surfaces $\mathcal{S}_{ICP}$ and $\mathcal{T}$ into volumes by calculating the SDF and the volumetric registration can be found by solving the following equation: 

\begin{equation}
    \hat{\textbf{T}}_\mu = \underset{\textbf{T}_\mu}{\arg\min} (\mathcal{C}(\textbf{T}_\mu; I_S, I_T)),
\end{equation}

\noindent where $\mathcal{C}$ is the cost function, $I_S$ is the source volume and $I_T$ is the target volume. 
$\textbf{T}_\mu$ is the transformation that transforms $I_S$ to $I_T$. 
We mask the volumes $I_S$ and $I_T$ to only include voxels that are less than 6 mm from the surface, since we are only interested in the 0-level isosurface and values far from this has limited influence on the final surface. 
The transformation is defined by the parameter vector $\mu$. 
We use a multi-level B-spline transformation with five resolution-levels. 
The cost function is a linear combination of the mean squared voxel-value difference similarity measure ($M S D$) and a bending energy penalty term ($\mathcal{P}_{BE}(\mu)$) as follows: 

\begin{equation}\label{eq:pointcor_cost}
    \mathcal{C} = \omega_1 MSD(\mu; I_S,I_T) + \omega_2 \mathcal{P}_{BE} (\mu).
\end{equation}

The weights ($\omega_1 = 1$, $\omega_2 = 1.5$) are chosen based on a grid search. 
We use stochastic gradient descent as the optimizer with 2048 random samples per iteration and a maximum of 500 iterations. 
The optimization is carried out using the Elastix library \cite{Klein2010elastix}\footnote{\url{https://elastix.lumc.nl/}}. 
The optimal transformation is applied to $\mathcal{S}_{ICP}$ and we denote the transformed surface $\mathcal{S}_{T}$. 

Since the volumetric registration is carried out on the SDF, there is no guarantee that the 0-level isosurface fits perfectly after the registration. 
As a final step, vertices in $\mathcal{S}_{T}$ is propagated to $\mathcal{T}$ using Markov Random Field regularization of the correspondence vector field \cite{Paulsen2003IPMI}. 

\subsection*{LAA decoupling} 
The LAA is decoupled from the LA with a plane through the LAA neck at its narrowest position.  
The right part of Figure \ref{fig:registration} shows the process. 
The points of the LAA neck were defined on the source LA mesh and the source mesh is registered to all target LAs in the dataset, to get an estimate of the neck points on the target mesh ($p_1, ..., p_m$). 
We aim to find a plane defined by a normal $n$ and a point $c$, with the optimal fit to the $m$ target neck points. 
The optimal plane can be found by solving: 

\begin{equation}
    \min_{c,||n||=1} \sum_{i = 1}^{m} ((p_i-c)^Tn)^2.
\end{equation}

Solving for $c$ we get the centroid of all points $c = 1/n\sum_{i=1}^m p_i$. 
We introduce the matrix $ A= [p_1-c, ..., p_m-c]$ and reformulate the minimization problem to:

\begin{equation} \label{eq:min2}
    \min_{||n||=1} ||A^Tn||_2^2.
\end{equation}

Equation \ref{eq:min2} can be solved using singular value decomposition (SVD) where $A = USV^T$. 
The plane is spanned by the first two vectors in $U$ and the plane normal can be directly read from the third vector $n = U(:,3)$.
The neck points ensures a coherent position of the cut-plane on all examples, but local adjustments to the individual shapes are necessary. 
Inspired by \cite{Leventic2019occluderplacement} we randomly add small perturbations to the plane normal and point, and choose the optimal separating plane as the plane slivering the LAA neck with the smallest cross sectional area. 
Examples of the placement of the decoupling plane can be seen in the supplementary material.

\subsection*{LAA morphology characterization}
We choose to refine the point correspondence on the decoupled LAAs, since the large differences in size and complexity between the LAA and the remaining LA makes it difficult to have good correspondence over the entire mesh. 
We therefore repeat the non-rigid volumetric registration with SDFs as described in Section \ref{sec:registration} for the decoupled LAAs.
We add five automatically derived landmarks on each LAA.
The first landmark is defined as the point on the LAA surface with the largest average Euclidean distance to all points along the decoupled edge. 
The remaining four landmarks are placed equidistantly around the decoupled edge, with the first landmark defined as the decoupled edge point closest to the first landmark. 
Examples of the landmarks can be seen in the supplementary material.
The landmarks serve as additional regularization of the non-rigid volumetric registration of the SDFs, where the term $\omega_3 \cdot \mathcal{P}_{CP} (\text{x},\text{y})$ is added to the cost function in Equation \ref{eq:pointcor_cost}.
Where $\omega_3  =0.15$ is found by experimental validation and $\mathcal{P}_{CP} (\text{x},\text{y})$ is a penalization of large Euclidean distance between corresponding landmarks. 

After point correspondence estimation, we can represent each LAA as a vector of concatenated (x, y, z) coordinates:  $ \mathbf{x}_i = [x_{i1}, y_{i1}, z_{i1}, \dots, x_{in}, y_{in}, z_{in}]^T$, $i=1,\ldots,s$, where $n$ is the number of vertices and $s$ is the number of shapes. 
We perform PCA on the shape matrix $\mathbf{D} = [( \mathbf{x}_1 - \overline{\mathbf{x}}) | \dots | ( \mathbf{x}_s - \overline{\mathbf{x}})]$, where $\overline{\mathbf{x}}$ is the average shape \cite{paulsen2002building}. 
This can be used to synthesize shapes within the learned shape space by adding linear combinations of the PCA loadings to the average shape:  $\mathbf{x}_\text{new} = \overline{\mathbf{x}} + \Phi\mathbf{b}$, where $\mathbf{b}$ is a vector of weights controlling the modes of shape variation and $\Phi = [\phi_1 | \phi_2 | \dots | \phi_t ]$ is the matrix of the first $t$ PCA loadings.

We use the PCA loadings to investigate whether distinct shape clusters are present, by fitting a number of multivariate Gaussian mixture models (GMMs) to the loadings. 
The probability density function of a GMM can be written as: 

\begin{equation} \label{eq:GMM}
    p(\text{x}) = \sum_{i=k}^{^K} \pi _k \mathcal{N} (\text{x} | \mu _k, \Sigma _k),
\end{equation}

\noindent where x is the PCA loadings, $\pi _k$ is the mixing coefficient, K is the number of mixture components and $\mathcal{N} (\text{x} | \mu _k, \Sigma _k)$ is the multivariate Gaussian distribution with mean $\mu_k$ and covariance matrix $\Sigma_k$.
To evaluate the fit of the GMM we compute the log-likelihood (LLH), which follows from Equation \ref{eq:GMM}:

\begin{equation}
    p(\text{x}|\pi, \mu, \Sigma ) = \sum_{i=1}^N \text{ln} \left( \sum_{k=1}^K \pi _k \mathcal{N} (\text{x} | \mu _k, \Sigma _k) \right).
\end{equation}

The number of shape clusters is determined using two-level cross-validation. 
The first level is a leave-one-out cross-validation. 
With N shapes in the data set, a model is trained on N-1 examples and the LLH is computed for the remaining shape. 
This is repeated until all N shapes have been used for testing once, and the average LLH is reported. 
The second level iterates through the $k=1,..,K$ mixture components. 
The number of shape clusters is equal to the number of mixture components with the highest mean test LLH.

    \subsection{Data}\label{sec:data}
\subsubsection*{Imaging Data Acquisition}
The data is acquired and provided by the Department of Radiology -- Rigshospitalet -- Copenhagen University Hospital. 
The data consists of 106 randomly selected participants, who had undergone a cardiac computed tomography angiography (CCTA) examination at Rigshospitalet in the period 2010-2013 for research purposes. 
Participant consent are obtained for all analyzed cases.
The CT-volumes are fully anonymized using the DICOM Anonymizer Tool provided by Medical Imaging Resource Center \footnote{Medical Imaging Resource Center, DICOM Anonymizer Tool, \url{mircwiki.rsna.org}}. 

\subsubsection*{Annotation procedure}
The LA and LAA were semi-manually annotated on the CT images by an expert annotator. 
A 3D U-net was trained on data from the 2017 Multi-Modal Whole Heart Segmentation (MM-WHS) challenge \cite{Zhuang2016,Zhuang2013} and predictions from this network provided an initialization of the LA segmentation on our dataset. 
The LAA was manually added and all segmentations checked and corrected using 3D slicer \footnote{3D Slicer, \url{www.slicer.org}} using 2D and 3D views. 
In total 90 CT images were manually annotated - 70 images were used for training the segmentation pipeline and 20 for testing. 
The remaining 16 images does not have a ground truth segmentation, but all 106 images were used for building the SSM.

\section{Results}
    \subsection{Segmentation and mesh model evaluation}\label{sec:results1}
We evaluate the segmentation based on Dice-score and a contour Dice-score that only captures errors in the vicinity of the segmentation boundary. 
The 3D mesh models is evaluated based on mesh regularity and similarity to ground truth using chamfer distance, earth movers distance, mesh accuracy, mesh completion and mesh cosine similarity \cite{Park2019}.
Additional evaluation details can be seen in \cite{Juhl2019}.

\begin{figure*}[!t]
\centering
\includegraphics[width = \linewidth]{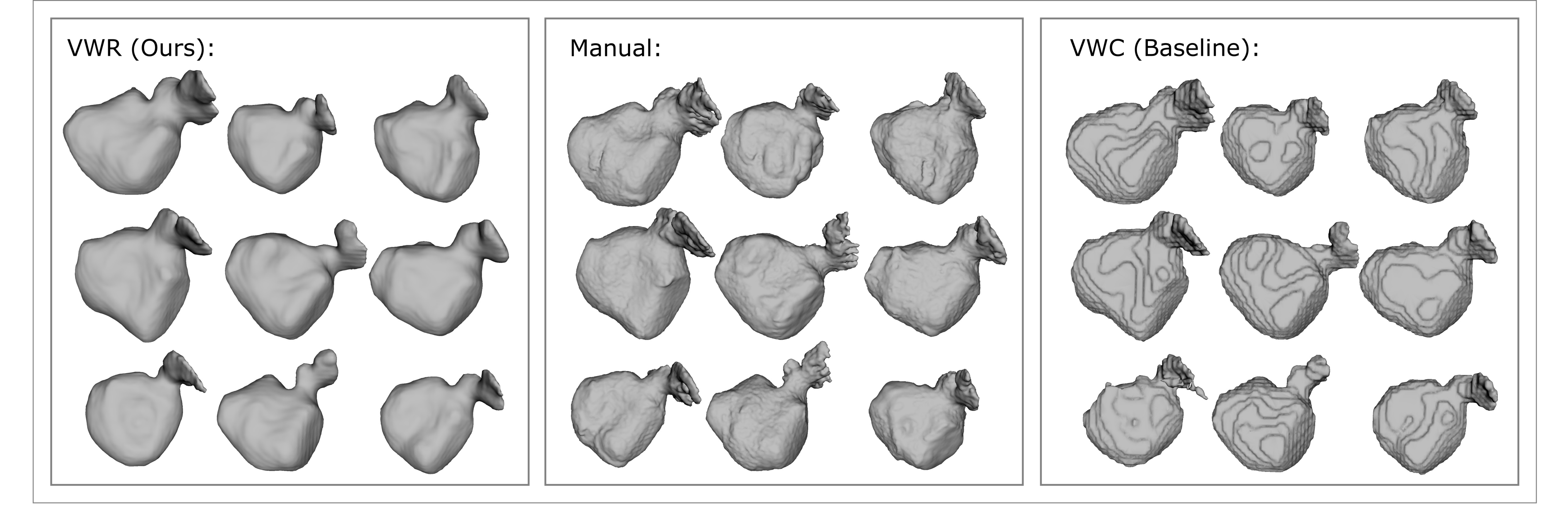}
\caption{Surface visualization of results from our voxel-wise-regression (VWR), manual expert annotation and voxel-wise-classification (VWC) without signed distance field regularization. The chosen examples correspond row-wise to the best, median and worst results evaluated based on dice-score. The baseline VWC with SDF regularization is visually equivalent to VWC without regularization and therefore omitted.}
\label{fig:LALAAvisualizations}
\end{figure*}

\begin{table}[htbp]
\caption{Evaluation results for voxel-wise-classification (VWC), signed distance field regularized VWC and voxel-wise-regression (VWR). \newline Arrow indicates whether higher ($\uparrow$) or lower ($\downarrow$) is better and the best result is highlighted in bold for each measure.}
\label{tab:results}
\begin{tabular}{p{2.8cm}p{1.5cm}p{1.5cm}p{1.5cm}p{1.5cm}p{1.5cm}p{1.5cm}} \toprule
                                   & \multicolumn{2}{c}{\textbf{Voxel-wise classification (VWC)}} & \multicolumn{2}{c}{\textbf{SDF regularized VWC}} & \multicolumn{2}{c}{\textbf{Voxel-wise regression (VWR)}} \\ 
                                   & Full            & LAA            & Full             & LAA              & Full            & LAA            \\\midrule
\textbf{$\uparrow$ Dice}           & 94.88           & 93.13          & 95.22            & 92.83            & \textbf{95.24}           & \textbf{93.26}          \\
\textbf{$\uparrow$ contour Dice}          & 74.42           & 79.20          & 75.14            & 79.24            & \textbf{75.79}           & \textbf{80.76}          \\
\textbf{$\downarrow$ Chamfer distance}& 0.915           & 0.844          & 0.877            & 0.888            & \textbf{0.779}           & \textbf{0.787}          \\
\textbf{$\downarrow$ EM distance}     & 2.813           & 1.920          & 2.972            & 1.907            & \textbf{2.475}           & \textbf{1.593}          \\
\textbf{$\downarrow$ Mesh Accuracy}    & 1.427           & 1.236          & \textbf{1.306}            & 1.338            & 1.310           & \textbf{0.964}          \\
\textbf{$\uparrow$  Mesh Completion}   & 0.801           & 0.789          & 0.805            & 0.756            & \textbf{0.818}           & \textbf{0.804}          \\
\textbf{$\uparrow$ Cosine Similarity}    & 0.915           & 0.867          & 0.917            & 0.860            & \textbf{0.951}           & \textbf{0.891}         \\\bottomrule
\end{tabular}
\end{table}

Table \ref{tab:results} shows the results of the quantitative evaluations. 
The Dice-score reveal that the segmentation methods produce good results with overlaps of $93 \%$ and $95 \%$ for the full LA and the LAA only.
The contour Dice-score increases with almost 2\% from the VWC to the VWR method, suggesting that an increased number of voxels is correctly classified at the vicinity of the boundary. 
We can see that the proposed VWR of SDFs produces mesh models that on average are less than 0.8 mm from the true mesh measured by the symmetric chamfer distance and 2.5 mm and 1.6 mm for the full mesh and LAA only, measured by the bifurcation from the Earth Movers distance. 
The mesh accuracy reveals that $90\%$ of the points on the full surface are within 1.3 mm from the true surface and in reverse $82\%$ of the true surface points are covered by the prediction within 5 mm. 
For the decoupled LAA $90\%$ of the predicted points are within 1 mm of the true points and $80\%$ of the true points are covered within 1 mm by the prediction. 
The cosine similarity is 0.951 for the full surface and 0.891 for the LAA only, which suggest that the normals on the true and predicted surface point in the same direction to a large extent. 
The normals affect the visual representation as seen in Figure \ref{fig:LALAAvisualizations}, where the difference in appearance between the proposed method and the baseline is clear. 
The VWR of SDFs produces smooth and continuous surfaces despite the extensive downsampling during processing, whereas the baseline VWC show severe staircase-like artefacts from the low processing resolution. 

\begin{figure}[!t]
\centering
\includegraphics[width = \linewidth]{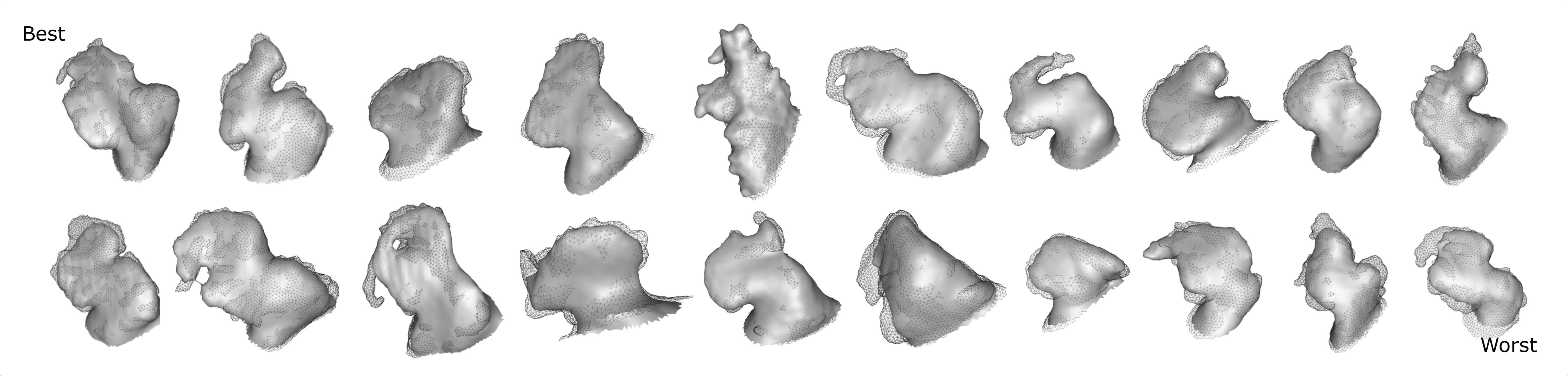}
\caption{Manual expert annotation (mesh) and our proposed method (full surface) from all 20 LAAs from the testset. The results are ranked based on Dice-score with the best result in the top left and the worst in bottom right.} 
\label{fig:LAAmeshresults}
\end{figure}

The average processing time from the original resolution input image to a triangulated mesh is 19.5 seconds.
The most time-consuming part is detecting the ROI since this requires loading a large image, up- and downsampling of input and output and detection of the largest connected component. 
These processing times is only slightly more than the fastest state-of-the-art methods and is low enough to make the proposed method usable in the clinic and for larger cohort studies. 

\subsection{Statistical shape model}\label{sec:results2}
\begin{figure}[b]
\centering
\includegraphics[width = .6\linewidth]{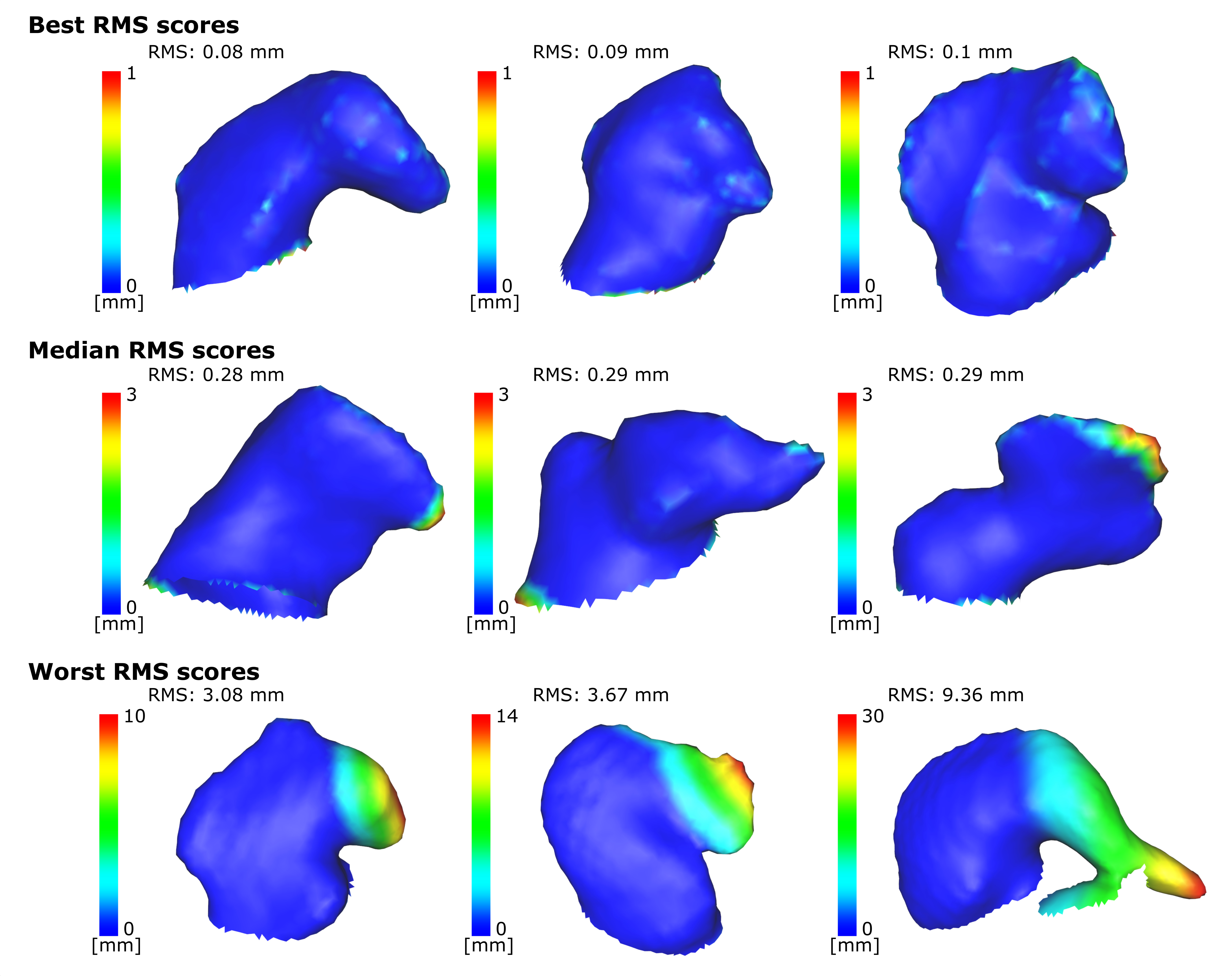}
\caption{Surface distance between the predicted LAA surface and the registered template. Row-wise, the examples correspond to the best, median and worst registrations based on root-mean-squared (RMS) distance.}
\label{fig:registration_results}
\end{figure}

The shape model is built on the automatically predicted LAA surfaces obtained using VWR of SDFs as described above on the full dataset of 106 patients. 
Figure \ref{fig:registration_results} shows the distance from the predicted surface to the surface in correspondence for the three best, median and worst registrations evaluated based on root-mean-square (RMS) distance between the two meshes. 
In the following, only registrations with an RMS-value below 2.5 were included in the shape model, which eliminates five examples.  

\begin{figure}[t]
\centering
\includegraphics[width = \linewidth]{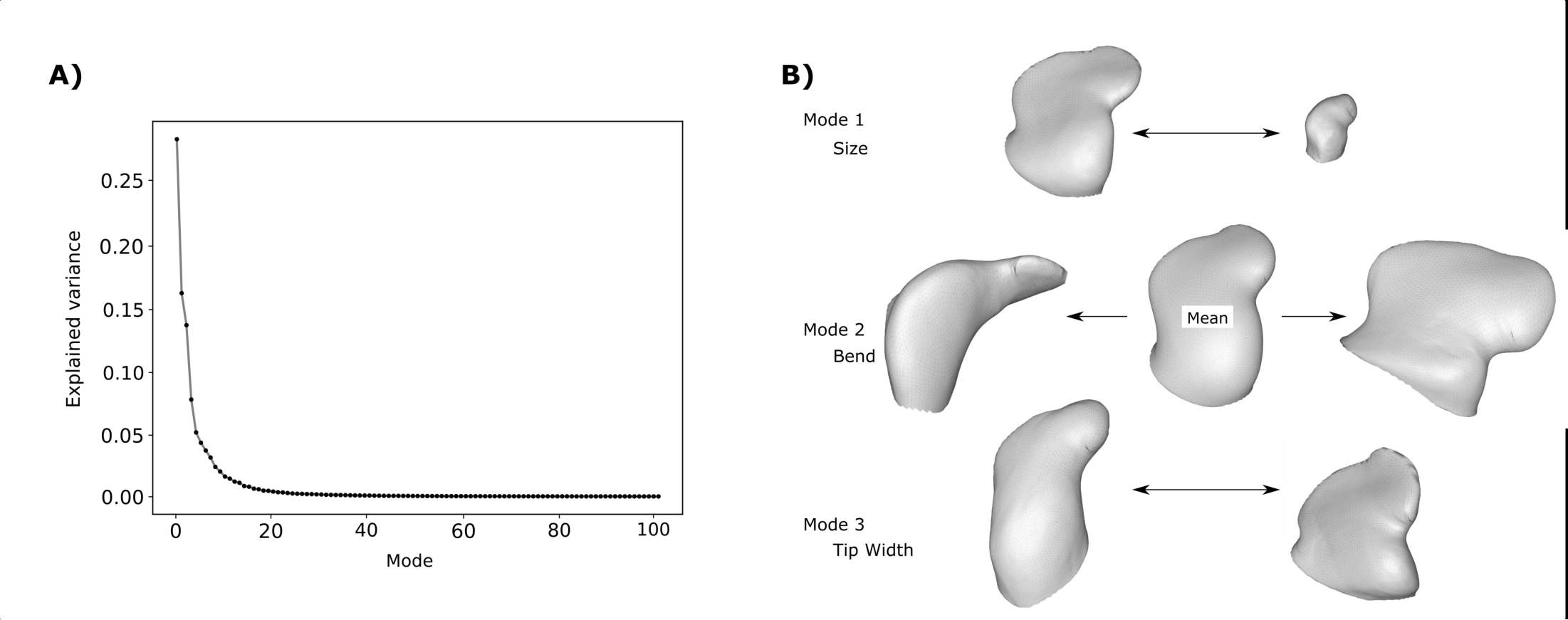}
\caption{A) Scree-plot showing the fraction of variation explained by each mode. 
B) Visualization of the average shape (center) together with the three main modes of variation with examples at $\pm 3$ standard deviations from the mean.
}
\label{fig:mainvar}
\end{figure}

Figure \ref{fig:mainvar}A shows the scree-plot from the PCA analysis.
We see that the first five PCA modes individually describe more than 95\% of the total variance, and that PCA modes above 20 explains only very little. 
We investigate the first three main modes of variation by setting each of the PCA loadings at the lower and upper limit ($\pm3$ standard deviations from the mean), while keeping the others at the average. 
The variations can be seen in Figure \ref{fig:mainvar}B. 
We observe that the first mode of variation mainly captures differences in LAA size. 
The secondary mode of variation seems to capture the bending of the LAA, whereas the third mode of variation describes the width of the LAA tip and the length of the main lobe.

\begin{figure}[b]
\centering
\includegraphics[width=\linewidth]{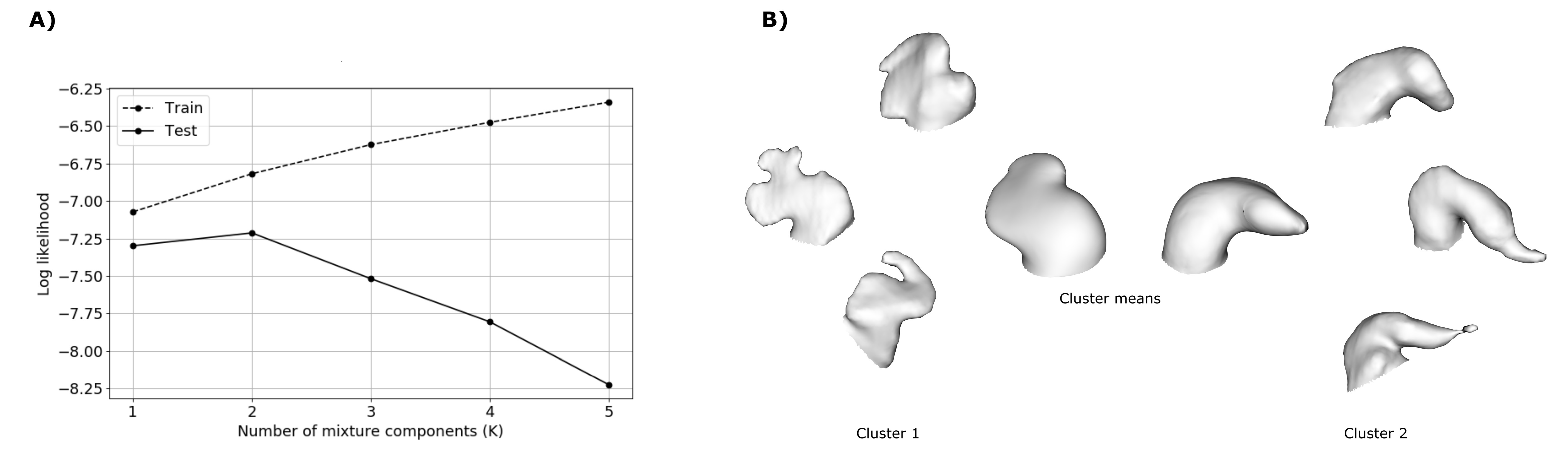}
\caption{A) Training and testing log-likelihood (LLH) after fitting gaussian mixture models (GMMs) to the first five principal components using two-level leave-one-out cross validation.
B) The cluster centers of the two Gaussian Mixture Model clusters are shown in the center and three examples of shapes belonging to each cluster is shown on either side.
}
\label{fig:cluster}
\end{figure}

Each LAA can thus be described as a linear combination of the variation from each of the PCA modes and new synthetic shapes can be generated by randomly assigning the weight of each variation \cite{Morales2019}.
One could assume that the entire PCA space would be valid LAA shapes, but it may also be that the LAA shapes lies in clusters, as it is assumed in the commonly used chicken-wing, cauliflower, windsock and cactus morphology groups \cite{wang2010left}.
We investigate this by fitting a number of multivariate GMMs to the PCA loadings and analyze the LLH for each of the fitted models as seen in Figure \ref{fig:cluster}A.
We use the first five modes of variation and find that the data most likely forms two clusters. 
A GMM with two clusters is fitted to the PCA-loadings of the LAA shapes in the data set. 
The two cluster centers are shown in Figure \ref{fig:cluster}B(center) and three examples of LAA anatomies belonging to each of the two clusters are shown on either side. 

\section{Discussion}

SDF-based processing has been widely used in computer vision and image processing. 
We show that SDFs can be regressed directly from CT images, and they can be used for segmentation and 3D mesh modelling of complex anatomies such as the LAA. 
Our method achieves better evaluations in all measures compared to the baseline VWC, while producing smooth and continuous surfaces. 
Considering the size of the test set, we considered a statistical test of significance inappropriate. 
We notice a general trend, where the baseline VWC network is slightly outperformed by the SDF regularized VWC, which again is slightly outperformed by our proposed method. 
This indicates that the regularization from the inversely weighted SDF on the loss function may guide the network towards making better decisions in this area. 
State-of-the-art segmentation methods (as described in the introduction) achieves Dice-scores around 90-95\% on their respective LAA-only datasets. 
Our segmentation results achieves an overlap of 93.26\% to the ground truth even though we process the full LA and LAA volume with relatively low memory requirements in a fully automatic pipeline. 
Furthermore, our proposed method is fast and can be run on readily available GPUs. 

Our method distinguishes itself from other methods using SDFs for medical image segmentation, in that our goal is to go beyond accurate voxel-labelling to produce accurate and high-quality meshes. 
\cite{dangi2019distance} formulate the SDF-regression as an auxilliary task during training, but it is removed at test-time. 
\cite{bui2019multi} uses the regressed SDF to refine the segmentation in a second network while \cite{Wang2017} uses an approximated heavyside function to create a differentiable mapping between the regressed SDF and the segmentation. 
They all conclude that the SDF contains valuable information, which improves the segmentation accuracy. 
\cite{Wickramasinghe2020} presented voxel2mesh, which also focus on creating high-quality meshes directly from images. 
They use a CNN based on the U-net architecture to extract feature vectors from the input image and a graph-based mesh decoder deforms the initial spherical mesh into the anatomy of interest. 
Compared to this, our proposed network model lies in the Euclidean space and is capable of representing any topology. 
In any case, graph-based deep learning is very promising and a potential future direction for our work. 

The surfaces obtained from the baseline VWC is found as the 0.5-level isosurface in the linearly upsampled probability map. 
This is done under the assumption, that the network is more uncertain in segmentation accuracy near anatomical border. 
In practice, this may often be true, but there is no proof that voxels close to the boundary have probabilities closer to 0.5. 
In the SDF framework the interpretation of the value in each voxel is however straight forward since it represents the approximated signed distance to the nearest surface. 
In order to improve on the mesh models from the baseline VWC method, one could make use of postprocessing such as surface smoothing. 
The necessary amount of smoothing to remove the staircase-like artefacts depend on the processing resolution and comes with a great risk of losing details of the complex LAA anatomy.
From the 3D mesh models in Figure \ref{fig:LALAAvisualizations} we observe that the surfaces obtained from our method is smooth compared to the manual segmentations. 
We argue, that the main reason for this is the downsampling of the input ROI to the processing size of $64^3$ voxels, which leads to the loss of smaller details on the LAA. 
It should however also be noted that the manual segmentations relies on image-intensities in the CT-image and for example the ruggedness of the LA wall is likely to be caused by noise in the image. 

In this paper, we have consciously avoided the term " LAA ostium" and instead used the term LAA decoupling. 
The rationale behind this is that the interpretation of the LAA ostium is different dependent on the reader. 
In this study we adopted the definition from \cite{Walker2012} and defined the ostium as the narrowest part of the LAA neck. 
This definition may not always be in line with the LAA ostium as defined during a LAA closure procedure. 
For building an LAA-only SSM it is important that the LAA is decoupled in a consistent way across all examples, whereas the exact position is considered less important. 
A thorough investigation of the definition of LAA ostium will therefore be left as a subject for further work. 


We have demonstrated that it is possible to build an SSM of the highly complex LAA anatomy.
The SSM relies on the estimated point correspondence between all LAA shapes. 
We have suggested one possible solution to this, but due to the large variations in shape, it is difficult to judge whether this correspondence is meaningful. 
Not all examples registered well to the common template, as shown in Figure \ref{fig:registration_results}. 
In two of these cases (bottom left and middle) the parts of the LAA with large distance between original and registered surface (Red and green area) can either be considered the tip of a bended LAA or a side lobe at the tip of a straight LAA. 
Our model aim for the latter interpretation, but does not have the necessary flexibility to fit the lobe without additional landmarks. 
Building the SSM also raises questions regarding the interpretation, such as how should we interpret the average LAA shape and how do we know if a combination of PCA modes leads to a plausible LAA shape? 
From the SSM presented in this paper, we have demonstrated that the LAA shape can be described with relatively few parameters, and that two clusters of LAA shapes exists, which may be described as chicken-wing and non-chicken-wing anatomy, with reference to the morphology classes from \cite{wang2010left}. 
The majority of the LAA shapes are classified as non-chicken-wing anatomies, which covers quite a large variety of shapes.  
We believe that this class can be further subdivided into more categories as more data becomes available.
The fully automatic pipeline we have presented in this paper may set the grounds for making larger datasets available in the future. 

\section{Conclusion}
We have shown that SDFs can be regressed directly from CT images and that the SDF is a versatile and accurate representation of shape. 
The SDF can be thresholded to construct a voxel-wise label map or the 0-level isosurface can be extracted to create a 3D mesh model of the LA and LAA anatomy. 
The segmentation results are on par with state-of-the-art with dice-score 95.24\% for the full LA and 93.26\% for the decoupled LAA.
The 3D mesh surfaces are on average less than 1 mm from the true surface and are smooth and continuous, which makes them usable for subsequent analysis, such as statistical shape modelling. 
We further demonstrate that SDFs can be utilized for registering 3D shapes to a common template and thereby establish point correspondence between the highly variable LAA anatomies. 
An SSM is built with registered shapes and the morphological variation is investigated.
We show that five PCA modes are sufficient to describe enough variation in shape to reveal two distinct clusters of LAA shapes, which may correspond to the chickenwing and non-chickenwing morphologies as known from clinical studies. 
The fully automatic pipeline derives user-independent morphological descriptors of shape, which opens new possibilities for larger studies on LAA morphology and its correlation to ischemic stroke.

\section*{Acknowledgments}
    \section*{Acknowledgements}
This work was supported by a PhD grant from the Technical University of Denmark - Department of Applied Mathematics and Computer Science (DTU Compute) and the Spanish Ministry of Science, Innovation and Universities under the Retos I+D Programme (RTI2018-101193-B-I00).
We thank Andreas Ellegaard for helpful advice on whole heart segmentation.

\bibliographystyle{unsrt}  
\bibliography{references}

\end{document}